\RequirePackage{fix-cm}
\documentclass[twocolumn]{svjour3}          
\smartqed  
\usepackage[colorlinks,linkcolor=red, citecolor=blue]{hyperref}
\usepackage{graphicx}
\usepackage{cite}
\usepackage{subcaption}
\usepackage{color}
\usepackage{amssymb}
\usepackage[ruled]{algorithm2e} 
\usepackage{url}
\usepackage{multirow}
\usepackage{indentfirst} 
\usepackage{natbib}
\begin{document}

\title{I3CL: Intra- and Inter-Instance Collaborative Learning for Arbitrary-shaped Scene Text Detection
}


\author{Bo Du \and Jian Ye \and Jing Zhang \and Juhua Liu* \and Dacheng Tao 
}



\institute{
B. Du and J. Ye are with National Engineering Research Center for Multimedia Software, Institute of Artificial Intelligence, School of Computer Science and Hubei Key Laboratory of Multimedia and Network Communication Engineering, Wuhan University, Wuhan, China. (e-mail: dubo@whu.edu.cn, leaf-yej@whu.edu.cn).\\
J. Zhang is with School of Computer Science, Faculty of Engineering, The University of Sydney, Sydney, Australia (e-mail: jing.zhang1@sydney.edu.au). \\
J. Liu is with Research Center for Graphic Communication, Printing and Packaging, and Institute of Artificial Intelligence, Wuhan University, Wuhan, China (e-mail: liujuhua@whu.edu.cn) (\emph{Corresponding author}).\\
D. Tao is with JD Explore Academy, China and School of Computer Science, Faculty of Engineering, The University of Sydney, Sydney, Australia (e-mail: dacheng.tao@gmail.com).\\
This work was done during Jian Ye's internship at JD Explore Academy.
}

\date{Received: date / Accepted: date}

\maketitle

\begin{abstract}

Existing methods for arbitrary-shaped text detection in natural scenes face two critical issues, $i.e.$, 1) fracture detections at the gaps in a text instance; and 2) inaccurate detections of arbitrary-shaped text instances with diverse background context. To address these issues, we propose a novel method named \textbf{I}ntra- and \textbf{I}nter-\textbf{I}nstance \textbf{C}ollaborative \textbf{L}earning (I3CL). Specifically, to address the first issue, we design an effective convolutional module with multiple receptive fields, which is able to collaboratively learn better character and gap feature representations at local and long ranges inside a text instance. To address the second issue, we devise an instance-based transformer module to exploit the dependencies between different text instances and a global context module to exploit the semantic context from the shared background, which are able to collaboratively learn more discriminative text feature representation. In this way, I3CL can effectively exploit the intra- and inter-instance dependencies together in a unified end-to-end trainable framework. Besides, to make full use of the unlabeled data, we design an effective semi-supervised learning method to leverage the pseudo labels via an ensemble strategy. Without bells and whistles, experimental results show that the proposed I3CL sets new state-of-the-art results on three challenging public benchmarks, $i.e.$, an F-measure of 77.5\% on ArT, 86.9\% on Total-Text, and 86.4\% on CTW-1500. Notably, our I3CL with the ResNeSt-101 backbone ranked the $1^{st}$ place on the ArT leaderboard. Code is available at \href{https://github.com/ViTAE-Transformer/ViTAE-Transformer-Scene-Text-Detection}{github.com/ViTAE-Transformer/ViTAE-Transformer-Scene-Text-Detection}.

\end{abstract}

\section{Introduction}
As a key procedure for text reading, scene text detection has gradually become an active topic in the computer vision community due to its wide range of applications \citep{zhang2020empowering}, such as autonomous driving, scene parsing, and visual-impaired navigation. Many excellent methods have been proposed recently thanks to the success of deep learning \citep{maskrcnn, east, pcr, fcenet, asts, recursive}. However, many issues in this task remain open and challenging, such as fracture detections at the gaps in a text instance and inaccurate detections of text instances with diverse background context, due to various factors including irregular shapes, complex fonts, and variable scales.

\begin{figure}[ht]
  \centering
  \includegraphics[width=1.0\linewidth]{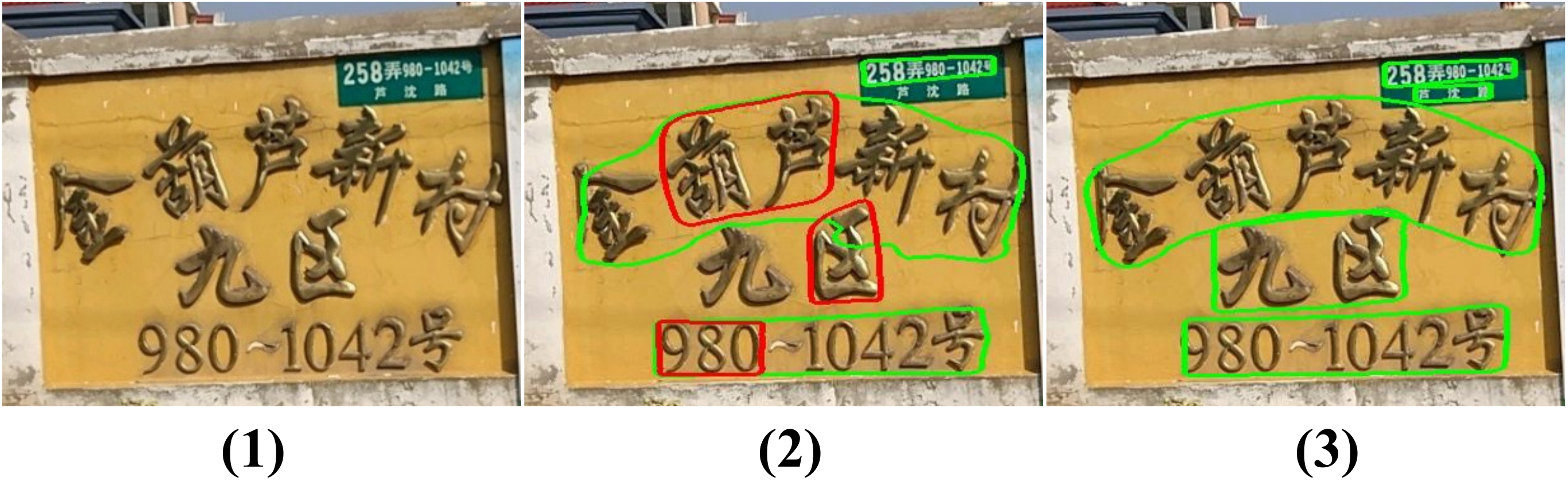}
  \caption{From left to right, a text image, the result of Mask R-CNN, and the result of our I3CL. Existing instance segmentation-based methods suffer from fracture detections due to the gaps inside the text (the bottom digital nameplate) and inaccurate detection due to the arbitrary shapes of different instances. Our I3CL produces much better results thanks to the intra- and inter-instance collaborative learning.}
  \label{fig1}
\end{figure}

\begin{figure}[ht]
  \centering
  \includegraphics[width=1.0\linewidth]{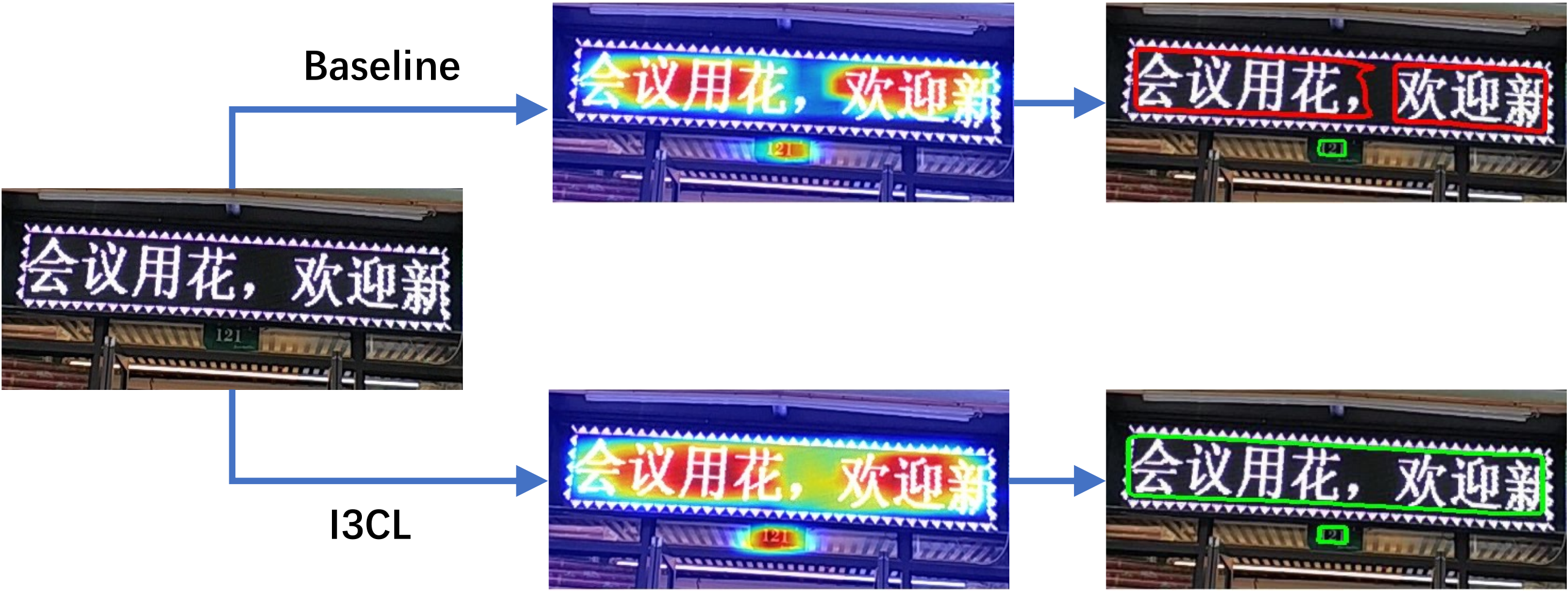}
  \caption{Existing Mask R-CNN based methods fail to detect the gaps inside the text and produce fracture detection while our I3CL model can address this issue through collaborative learning.}
  \label{fig2}
\end{figure}

Most of the previous methods \citep{east, textboxes, seglink} are designed for horizontal or multi-oriented text detection and have encountered troubles in dealing with arbitrary-shaped text. Some methods propose to represent curved text with a set of characters, which is time-consuming and requires complex post-processing. In recent studies, inspired by Mask R-CNN \citep{maskrcnn}, instance segmentation-based approach is proposed to address the problem of detecting text of various shapes. Nevertheless, simply applying Mask R-CNN to scene text detection also has some thorny problems.

As illustrated in Figure \ref{fig2}, one of the main problems is the fracture detection at the gaps in a text instance. When detecting text with extremely scattered and misaligned characters, the detection model may produce low text feature responses in the regions of gaps between characters because of its weak text feature representation capacity in these regions. As a result, the text detector will suffer from fracture detections. Therefore, how to learn a strong text feature representation for both characters and gaps in the text instance matters for improving the detection performance. Besides, another problem is the inaccurate detection of text instances due to diverse background, such as false positives, missed detections, as well as incomplete contours. Although existing methods learn to detect all text instances within an image through end-to-end modeling, they treat them as individual instances during training. Consequently, existing methods have difficulties in distinguishing texts from the complex background and are prone to generate inaccurate detection results. In this paper, we argue that the text instances within an image probably have some kind of commonness. It refers to the common properties between different text instances due to similar font, color, size, and shared background context, which represent the semantic information of text instances and are completely different from the background semantic. Similar to the term of long-range dependencies between pixels within texture regions or objects in context, we term the relationship between text instances sharing the commonness as the long-range dependencies. How to exploit the dependencies between text instances and leverage the global context from the same background matters for learning a strong text feature representation.

To address these issues, we proposed a novel scene text detector based on Intra- and Inter-Instance Collaborative Learning (I3CL), which can effectively detect arbitrary-shaped scene texts. On the one hand, we first observe that the gaps in a text contain useful semantic information distinct from the background, since they are connected to the characters on both sides. We suspect that existing methods have limited performance because they are trapped by the limited receptive fields and thus have weak representation capacity for these gap regions. Based on the observation, we propose an intra-instance collaborative learning module, which treats a text as a combination of characters and gaps and learns discriminative features for them. Specifically, it consists of a cascade of three convolutional blocks, in each of which we use two convolutional layers with asymmetric horizontal and vertical convolutional kernels, and a convolutional layer with a regular convolutional kernel in parallel to them. In this way, it can model both character and gap regions in multi-oriented texts via an ensemble of paths with different receptive fields. On the other hand, to exploit the dependencies between different text instances, we propose an inter-instance collaborative learning module based on an instance-based transformer structure and a global context module, where the texture instance features are used as a token sequence to model the dependencies while the global context from the same background will be learned to supplement the above text features. By integrating these modules into a unified end-to-end trainable network, I3CL can learn a more discriminative feature representation for arbitrary-shaped scene text detection. In addition, to use unlabeled data to improve the performance, we design a simple yet effective pseudo label generation method based on an ensemble strategy, which can mitigate the problems of missed and false detections when producing reliable pseudo labels. 

The contribution of this work is four-fold. Firstly, we devise an intra-instance collaborative learning module to learn a unified feature representation for both character and gap regions in the text instance. Secondly, we devise an inter-instance collaborative learning module to exploit the dependencies between text instances within an image. Thirdly, we propose a pseudo label generation method based on an ensemble strategy to harvest the unlabeled data in a semi-supervised learning (SSL) framework. Finally, Our I3CL model outperforms existing methods and sets new state-of-the-art results on three challenging public benchmarks.

\section{Related Work}
\subsection{Scene Text Detection}

\textbf{Regression-based methods} follow the generic object detection framework and localize texts by directly regressing the bounding boxes of text instances. For example, EAST \citep{east} used efficient pixel-level regression for text objects without using the anchor mechanism and proposal generation. Based on SSD \citep{ssd}, TextBoxes \citep{textboxes} modified the aspect ratio of anchors and added a text-box layer using a horizontal convolutional kernel. Further, TextBoxes++ \citep{textboxes++} applied quadrilaterals regression for multi-oriented text instances. SegLink \citep{seglink} proposed to employ fully convolutional networks to detect text segments and model their link relationships. DGGR \citep{dggrn} first used a graph convolutional network to model relational reasoning of text components, and then grouped them into text results by linking merging. Although these methods have achieved good performance for quadrilateral text detection, most of them can not handle irregular shaped texts well due to the limited geometric representation ability.

\textbf{Segmentation-based methods} can accurately describe scene texts in various shapes using pixel-level segmentation masks. For example, TextSnake \citep{textsnake} proposed a flexible and general text representation for arbitrary-shaped texts by predicting the text center line and text regions with geometry attributes. PSENet \citep{psenet} generated whole text boundary by performing progressive scale expansion of text regions using different scale kernels. Inspired by Mask R-CNN \citep{maskrcnn}, SPCNet \citep{spcnet} proposed a supervised pyramid context network to detect arbitrary-shaped texts based on instance segmentation. CRAFT \citep{craft} detected the text by clustering characters boxes according to exploring affinity between characters. For real-time detection, DB \citep{db} designed a differentiable binarization module to perform the binarization process in a segmentation network. TextFuseNet \citep{textfusenet} adopt a multi-path fusion architecture to fuse three levels of features for text detection. However, these methods still suffer from fracture detections and inaccurate detections. Moreover, these methods treat each text instance as an individual object for learning and training, and pays no attention to the adverse influence caused by gaps in a text, which make them suffer from fracture detections and inaccurate detections. In contrast to them, we propose a novel idea of intra- and inter-instance collaborative learning to learn better feature representation by exploiting the intra-instance characteristics and inter-instance dependencies.

\subsection{Collaborative Learning}
Collaborative learning (CL) has been widely used in different visual tasks. For example, Wang $et$ $al$. \citep{wang2018collaborative} proposed a collaborative learning framework of object detection by enforcing partial feature sharing and prediction consistency to train a weakly supervised learner and a strongly supervised learner jointly. CDCL \citep{cdcl} presented a Cross-Dataset Collaborative Learning method to improve the generalization and discrimination of feature representations for semantic segmentation. Song $et$ $al$. \citep{song} introduced a collaborative learning network where multiple classifier heads of the same network are simultaneously trained to improve generalization and robustness to label noise.  Zhang et al.\citep{cln} proposed to improve text detection via collaborative training of weakly supervised text classification network and supervised text detection network. In the context of scene text detection, existing methods pay little attention to the gaps in the text and handle text instances separately, resulting in a weak text feature representation ability. By contrast, we propose a novel collaborative learning model to learn a unified feature representation for both characters and gaps in the text and exploit the dependencies between different instances, which is an instance-level collaborative learning different from the task-level collaborative learning in \citep{cln}.

\begin{figure*}[ht]
  \centering
  \includegraphics[width=1.0\linewidth]{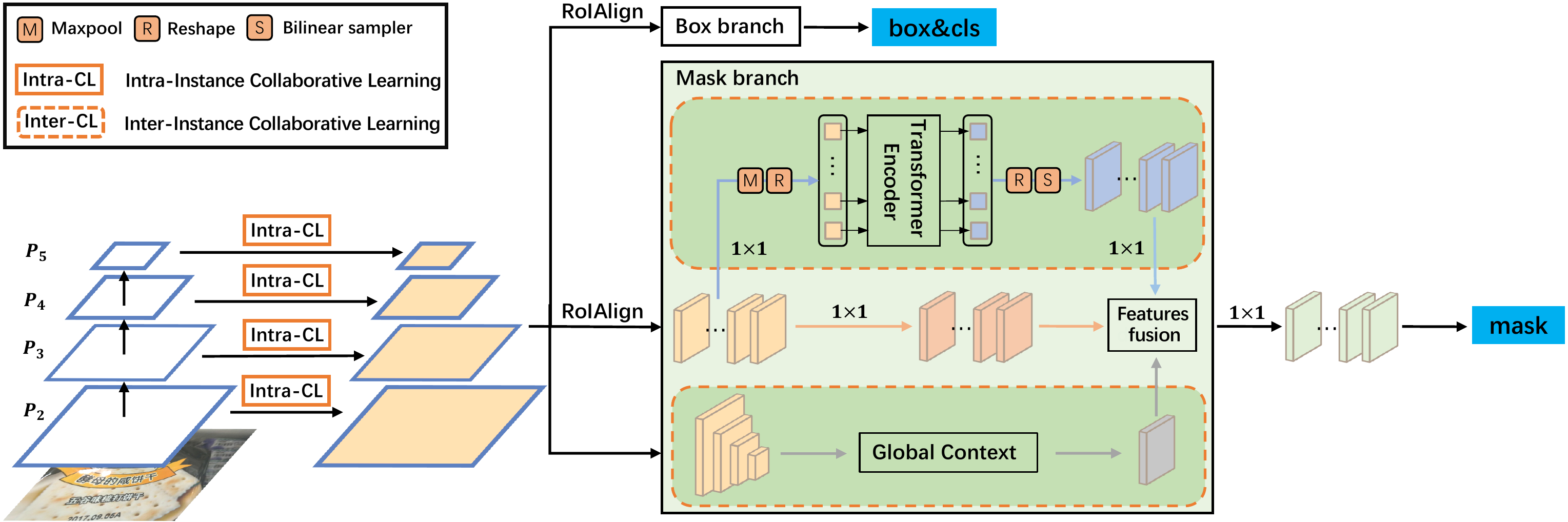}
  \caption{The overall pipeline of the proposed I3CL. Based on the Mask R-CNN, it refines the feature map at each scale of the feature pyramid via the Intra-Instance Collaborative Learning module, and further embeds the features of text instances and global context using the Inter-Instance Collaborative Learning module in the mask branch.}
  \label{fig3}
\end{figure*}

\subsection{Self-training with Pseudo Labels}
Self-training using pseudo labels is a learning paradi-gm associated with constructing models in semi-superv-ised learning, which leverages the model’s own confident predictions to produce the pseudo labels for unlabeled data \citep{noisestudent,cag,zhang2021towards}. Xie $et$ $al$. \citep{noisestudent} proposed a Noise Student method inspired by knowledge distillation with equal-or-larger student models. Zhang $et$ $al$. \citep{cag} proposed to use the category centers of the source domain features as guiding anchors, which can be used to determine the active features of the target domain and generate pseudo labels for semantic segmentation. Zou $et$ $al$. \citep{crst} proposed the confidence regularized self-training to avoid putting overconfident pseudo labels on wrong classes, which may leading to deviated solutions with propagated errors. Zhang $et$ $al$. \citep{prototypical} proposed to use the feature distances from prototypes to estimate the likelihood of pseudo labels to facilitate online correction in the course of training. Yang $et$ $al$. \citep{interactive} designed multiple detection heads that predict pseudo labels for each other to provide complementary information. Unlike the above methods that may suffer from the erroneous pseudo labels from a single model, we proposed a new pseudo labeling method based on an ensemble strategy to produce reliable pseudo labels for text detection.

\section{Methodology}

\subsection{Overview}
The overall framework of the proposed I3CL model is illustrated in Figure \ref{fig3}. As shown, the basis of I3CL pipeline is built upon the Mask R-CNN framework. Firstly, the input text image is fed into the backbone network with an FPN \citep{fpn} architecture to generate a multi-scale feature pyramid, denoted as \{$P_{2}$, $P_{3}$, $P_{4}$, $P_{5}$\}, which have the same size as the input image with the down-sampling factors of \{4, 8, 16, 32\}, respectively. Secondly, the Intra-Instance Collaborative Learning (Intra-CL) module is used to further refine the text fracture of both characters and gaps implicitly on the feature maps at each scale of the feature pyramid. The detailed network structure of Intra-CL will be presented in Section~\ref{subsec:intraCL}. Next, we use a region proposal network (RPN) to produce text proposals for subsequent procedures. After that, box regression and mask prediction are carried out in two parallel branches. The box branch further refines and classifies text proposals. In the mask branch, we devise an Inter-Instance Collaborative Learning (Inter-CL) module to perform collaborative learning among all positive text instances. The text features from the Inter-CL module and the original ROIAlign module are fused into more discriminative features and used in instance segmentation to generate precise text contours. The detailed network structure of Inter-CL will be described in Section~\ref{subsec:interCL}. Besides, a pseudo label generation method based on ensemble strategy for semi-supervised learning will be introduced in Section~\ref{subsec:semisupervised} in detail. 

\begin{figure}[t]
  \centering
  \includegraphics[width=\linewidth]{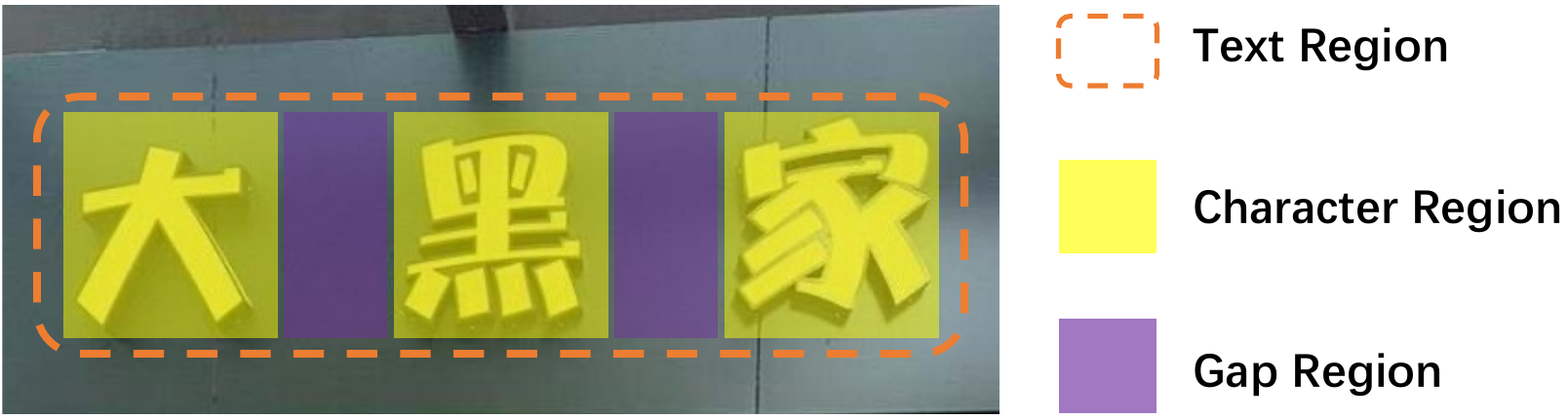}
  \caption{The text region consists of spaced character regions and gap regions. Compared with the background, the gap regions are surrounded by characters in both sides, and contain rich text-related information in a long range. Therefore, it requires exploiting long-range dependencies between characters, between gaps, as well as between characters and gaps to learn a complete representation for the whole text instance. }
  \label{fig4}
\end{figure}

\subsection{Intra-Instance Collaborative Learning}
\label{subsec:intraCL}
Existing methods focus on learning discriminative features for the character regions in the text instance while paying little attention to the gap regions between the characters, which may result in fracture detections ($i.e.$, false positives detection) at the gaps due to a weak feature representation ability. In this paper, we treat the text instance as a spaced combination of both characters and gaps, as illustrated in Figure \ref{fig4}. In other words, the characters are spaced by the gaps while the gaps are also surrounded by characters on both sides, indicating that there are long-range dependencies between characters, between gaps, as well as between characters and gaps. To exploit the dependencies and learn a unified discriminative feature representation for both characters and gaps, we propose the Intra-CL module consists of a cascade of three convolutional blocks with multiple receptive fields.

\begin{figure}[t]
  \centering
  \includegraphics[width=0.8\linewidth]{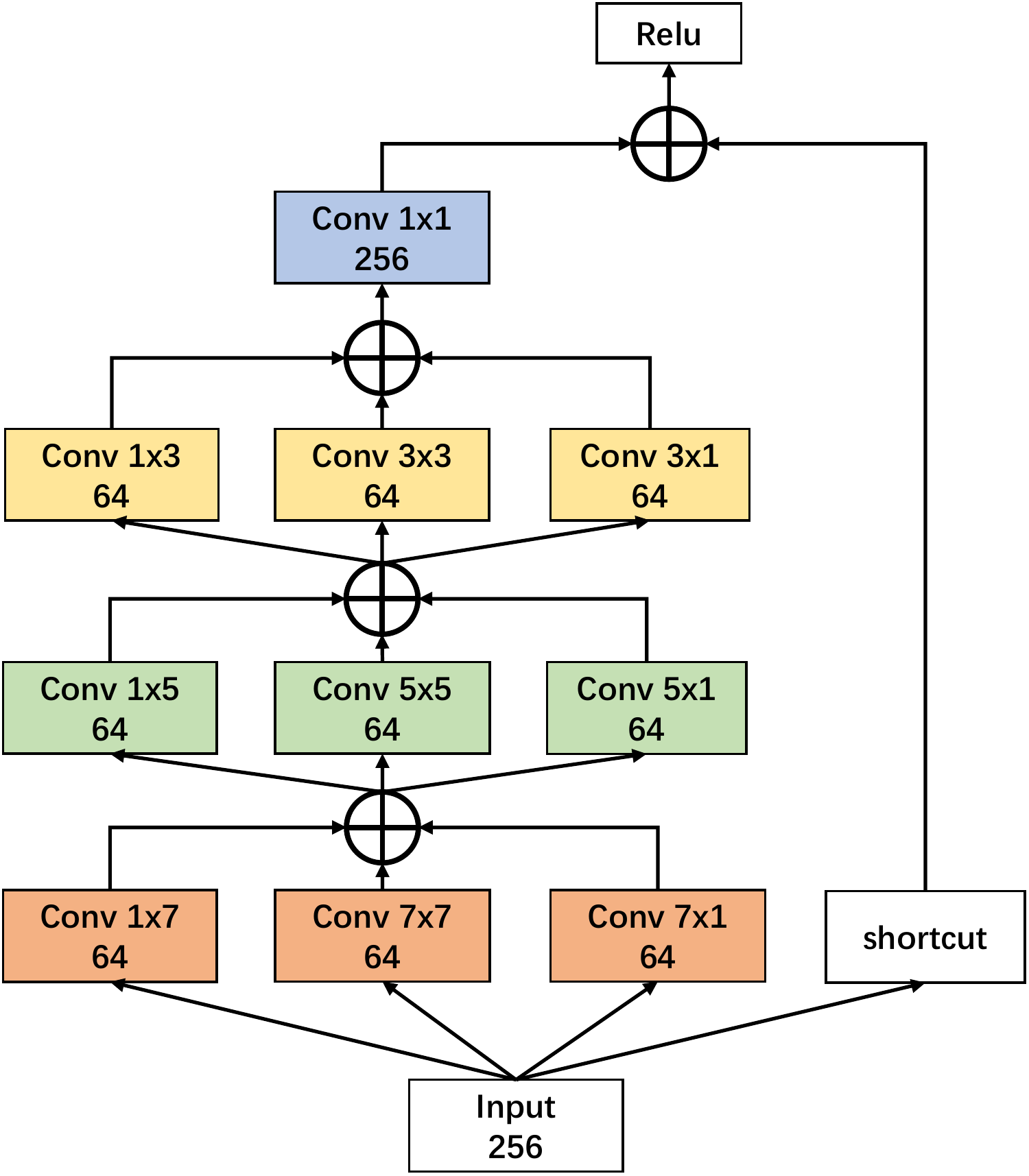}
  \caption{The architecture of the Intra-Instance Collaborative Learning module. It consists of a cascade of three convolutional blocks, each of which contains two convolutional layers with asymmetric horizontal and vertical convolutional kernels, $i.e.$,  $k\times1$ and $1\times k$, and a convolutional layer with a regular $k\times k$ convolutional kernel in parallel to them. In this way, it can model both character and gap regions in multi-oriented texts via an ensemble of paths with different receptive fields.}
  \label{fig5}
\end{figure}

As shown in Figure \ref{fig5}, the Intra-CL module is composed of a cascade of three convolutional blocks. Each block contains three parallel convolutional layers with asymmetric horizontal and vertical convolutional kernels, $i.e.$,  $k\times1$ and $1\times k$, as well as a regular $k\times k$ convolutional kernel. In our work, $k$ is set to 7, 5, and 3, respectively. The features from the three layers are summed and fed into the subsequent block. In this way, the Intra-CL module indeed contains an ensemble of paths with multiple receptive fields. We also add a residual connection between the input feature and the fused feature of the last block since it has been proved that learning with residual connections is much easier and converges faster. It is noteworthy that we use asymmetric kernels to enable the Intra-CL module adapt to multi-oriented texts. Besides, we employ a large kernel at the first block and smaller ones in the subsequent blocks because the Intra-CL module is expected to learn long-range dependencies between characters and gaps at first and then gradually focus on the central region of either the character or gap to learn a discriminative feature representation. An ablation study of the design of the Intra-CL module is conducted in Section~\ref{subsec:ablation}.

Unlike the inception-like modules in IncepText \citep{inceptext} enlarging received field for horizontal text detection by stacking separable convolutions ($1\times k$ and $k\times 1$) sequentially and convolution layers with different kernels in parallel, I3CL aims to learn text representations in longer ranges for arbitrary-shaped text detection. To this end, first, we utilize separable convolutions in parallel and stack convolution layers with different kernels sequentially. Second, we use large kernels at the shallow layers and small kernels at the deep layers to make the network gradually focus on the central region of either the character or gap to learn a complete text representation. Third, the parallel and serial structure in Intra-CL implies an ensemble of $3\times 3 \times3=27$ paths, each of which corresponds to a unique combination of convolution kernels and has a specific receptive field.

By deploying the proposed Intra-CL module at each scale of the feature pyramid, our detection model can exploit the long-range dependencies between characters and gaps in text region through the information flows among different paths and implicitly learn a unified feature representation for both characters and gaps, therefore effectively mitigating the fracture detection issue due to the gaps in a text instance.

\subsection{Inter-Instance Collaborative Learning}
\label{subsec:interCL}
Following Mask R-CNN \citep{maskrcnn}, we use the RoIAlign to extract the RoI features of size $H$ $\times$ $W$ $\times$ $C$ from the multi-scale feature pyramid for $M$ positives proposals, which will be fed into the box branch and mask branch, respectively. To model the dependencies between text instances, we applied the transformer structure in Inter-CL module as shown in Figure~\ref{fig3}. Firstly, the $M$ RoI features are fed into a 1 $\times$ 1 convolution layer to reduce their channel dimension from $C$ to $C_{0}$. Then, their spatial resolution is also reduced from $H$ $\times$ $W$ to $h$ $\times$ $w$ by using Adaptive Max-Pooling. Next, we flatten each feature into a vector of size $1 \times (h \times w \times C_{0})$. In this way, we obtain a sequence of $M$ token features (denoting as $q$), whose feature dimension has been significantly reduced. The sequence $q$ is fed into a transformer encoder, which has three regular encoder layers with four heads of self-attention layers. And the output feature dimension of the feed-forward network in the transformer is $h \times w \times C_{0}$. Via the multi-head attention module, the long-range dependencies between different text instances in an image can be captured by adaptively attending to specific text instances that have similar background context or font appearance for any text instance. In this collaborative learning way, the representation ability of learned features can be improved. Afterward, the sequence $q$ will be reshaped to a set of enhanced 2D visual features of size $h$ $\times$ $w$ $\times$ $C_{0}$, which will be upsampled using bilinear interpolation and transformed using a $1 \times 1$ convolution layer to recover the feature dimension as $H$ $\times$ $W$ $\times$ $C$. In this paper, the typical setting of the aforementioned parameters are $C$=256, $C_{0}$=32, $H$=$W$=14, $h$=$w$=3, and $M$ is the number of positive instance proposals. The whole process can be described as follows:
\begin{equation}
\quad\quad \textit{q} = \textit{Reshape}(\textit{AdaptiveMaxpool}(\textit{Conv}_{1\times1}(\textit{f}))),
\label{equ1}
\end{equation}
\begin{equation}
\quad\quad \textit{q}^{\textit{TE}} = \textit{TransformerEncoder}(\textit{q}),
\label{equ2}
\end{equation}
\begin{equation}
\quad\quad \textit{q}^{*} = \textit{Conv}_{1\times1}(\textit{BilinearInterp}(\textit{Reshape}(\textit{q}^{\textit{TE}}))),
\label{equ3}
\end{equation}
where $f$ denotes the RoI features of $M$ text instances, $q^{TE}$ denotes the learned features by the transformer encoder, and $q^{*}$ is the recovered 2D visual features. 

 \begin{figure}[ht]
  \centering
  \includegraphics[width=0.75\linewidth,height=0.8\linewidth]{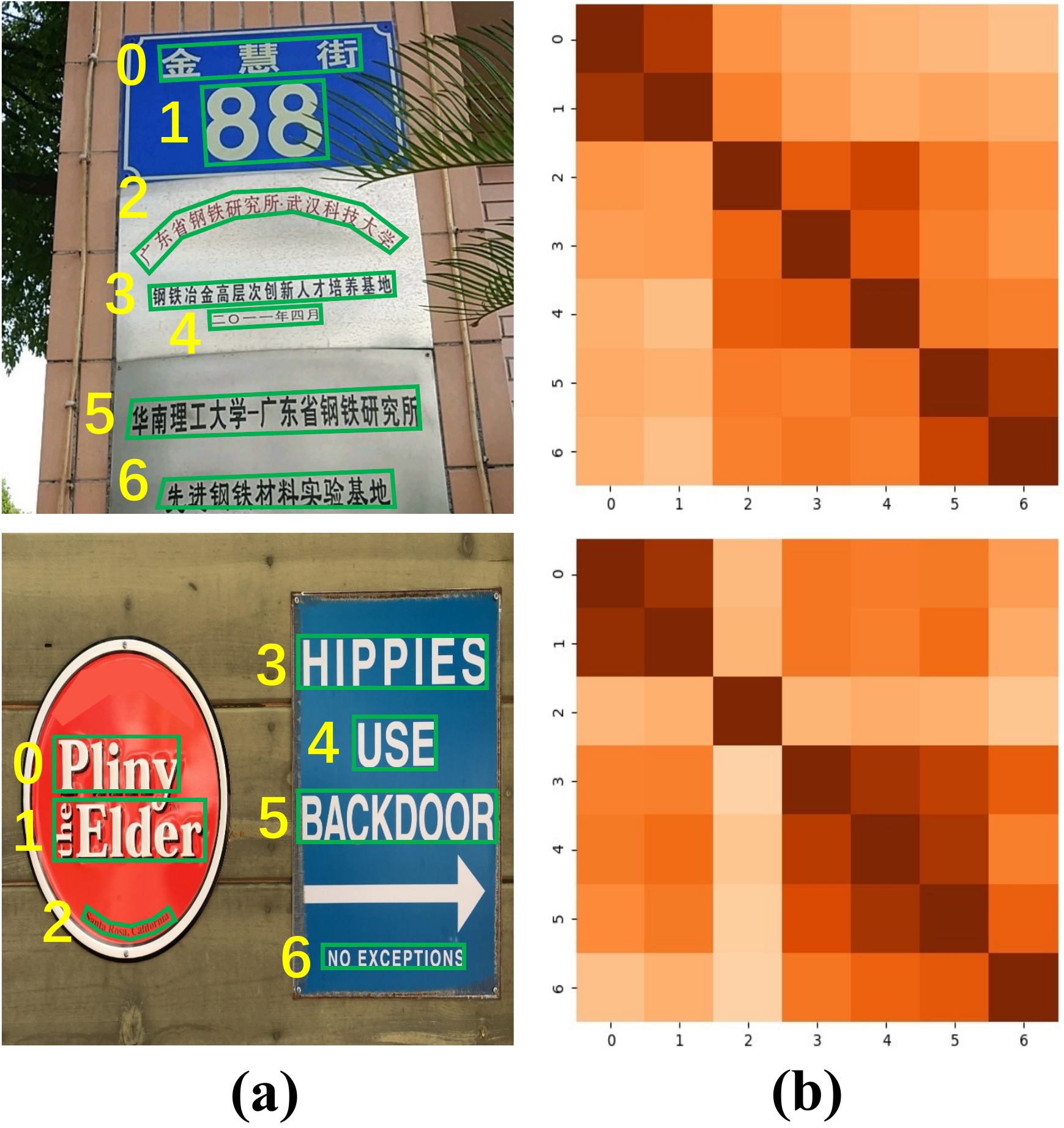}
  \caption{(a) Different text instances on an image. (b) Attention map of the dependencies between text instances. The darker the color, the closer the dependency between two instances. As shown, the dependencies between different text instances are influenced by background context, font, and scale, etc.}
  \label{fig6}
\end{figure}

\begin{figure}[t]
  \centering
  \includegraphics[width=0.9\linewidth]{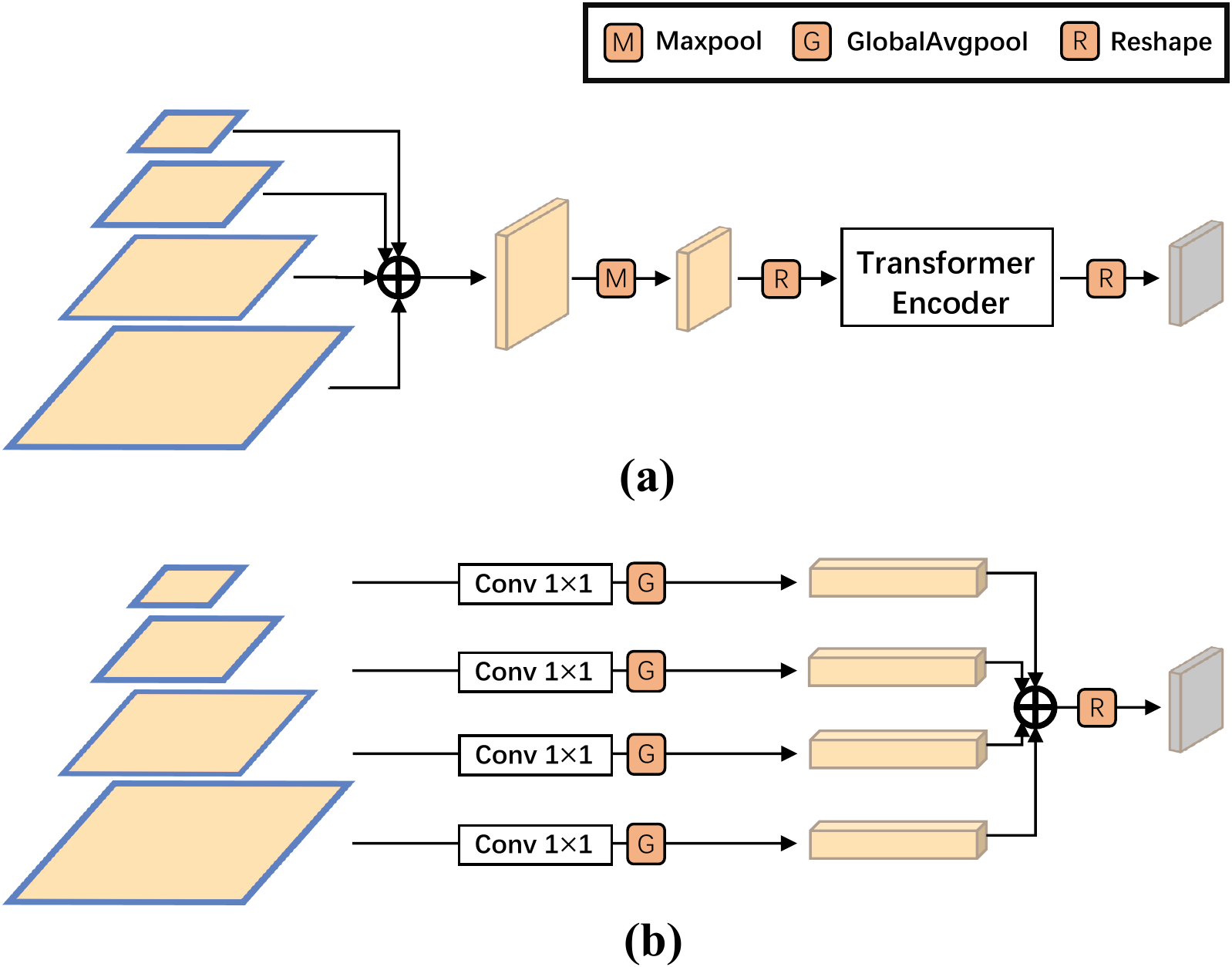}
  \caption{(a) Global context extraction structure based on a Transformer encoder. (b) Global context extraction structure based on global average pooling.}
  \label{fig7}
\end{figure}

Existing methods typically detect texts according to the RoI feature that hardly pay attention to the global context from shared background for text instances with-in an image, which may tend to produce inaccurate detection results. We introduce two different structure designs in Inter-CL module to extract global context, which can be seen in Figure \ref{fig7}. \textbf{1)} The first structure is built upon a pixel-based transformer. After obtaining the unified representation from all the levels of the feature pyramid like in TextFuseNet \citep{textfusenet}, we flatten the feature maps into a sequence of tokens, where each token is a feature vector at a specific pixel position on the feature maps as shown in Figure \ref{fig7}(a). In this way, we extract the global context by modeling the long-range dependencies between different pixels on the feature maps. \textbf{2)} For the second structure shown in Figure \ref{fig7}(b), each level of feature pyramid is aggregated into a global context vector by a $1 \times 1$ convolution layer and global average pooling, and then we fuse global context from different scales through element-wise summation. The global context will be fused with the original RoI features and the enhanced RoI features from the transformer encoder via element-wise summation to generate the discriminative text representation for text instance segmentation as shown in Figure \ref{fig3}.

As we have discussed, the text instances within an image probably have strong dependencies since they may share a same background or have a same font style, color, and scale, as illustrated in Figure \ref{fig6}(a). Based on the Inter-CL module, we can effectively model dependencies between text instances via the self-attention mechanism as demonstrated in Figure \ref{fig6}(b), which is beneficial for learning discriminative feature representation. It is noteworthy that we have not utilized the Inter-CL module in the box branch for the following three reasons. First, there are both positive and negative text samples in the box branch, while we only need to model the dependencies between different positive text instances rather than those negative ones, which are primarily used for training the classifier. Second, there are a lot of negative samples which will result in a long sequence and a bulk of computations if we directly apply the Inter-CL module based on them. Third, since we derive the detection results from the predicted masks rather than the quadrilateral bounding boxes for the arbitrary-shaped scene texts, therefore we only deploy the Inter-CL module in the mask branch. 

\begin{figure*}[ht]
  \centering
  \includegraphics[width=1\linewidth]{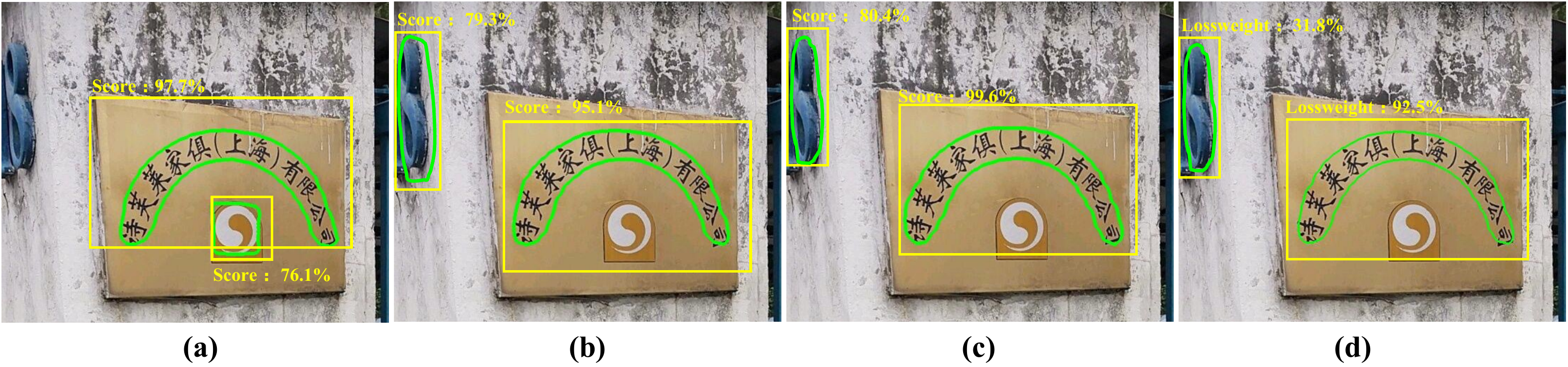}
  \caption{(a)-(c) are the detection results from different teacher models, and (d) is the generated reliable pseudo label based on above results. }
  \label{fig8}
\end{figure*}

\subsection{Semi-supervised Learning}
\label{subsec:semisupervised}
SSL has been widely applied in various deep learning tasks, which can effectively use unlabeled data to improve performance. Among them, self-training based on pseudo labels is one of the most common methods in SSL. However, existing methods obtain pseudo labels from the detection results only via a customized confidence threshold, ignoring the errors of missed and false detections in object regression. To mitigate the side effect of the problem, we propose a more reliable pseudo label generation method as described below.

\begin{algorithm}
\caption{Pseudo-label Generation} 
\label{algorithm1}
    \KwIn{$Det_{A}=\{text_{i}=(m_{i},b_{i},s_{i})\}^{I}_{i=0}$,
           $Det_{B}=\{text_{j}=(m_{j},b_{j},s_{j})\}^{J}_{j=0}$,
           $Det_{C}=\{text_{k}=(m_{k},b_{k},s_{k})\}^{K}_{k=0}$}
    \KwOut{$L_{p}=\{text_{q}=(m_{q},b_{q},w_{q})\}^{Q}_{q=0}$}
    \Begin{
    $ L_{p} = \{\}$;\\
    \For{$text_{i} \in Det_{A}$}{
    
    \If{$\exists~text_{j} \in Det_{B}$ \rm{\textbf{and}} $text_{k} \in Det_{C}$, 
        $iou_{ij} > T$ \rm{\textbf{and}} $iou_{ik} > T$}{
        $m_{q}$ = \textbf{Overlap-Mask}($m_{i}$,$m_{j}$,$m_{k}$)\\
        $b_{q}$ = \textbf{Soft-Box}($b_{i}$,$b_{j}$,$b_{k}$)\\
        $w_{q}$ = $s_{i}*s_{j}*s_{k}$\\
        $L_{p} = L_{p}\bigcup (m_{q},b_{q},w_{q})$
        }
    
    \ElseIf{$\exists~text_{j} \in Det_{B}$ \rm{\textbf{and}} $iou_{ij} > T$}{
        $m_{q}$ = \textbf{Overlap-Mask}($m_{i}$,$m_{j}$)\\
        $b_{q}$ = \textbf{Soft-Box}($b_{i}$,$b_{j}$)\\
        $w_{q}$ = $s_{i}*s_{j}*\alpha$\\
        $L_{p} = L_{p}\bigcup (m_{q},b_{q},w_{q})$
        }
    
    \ElseIf{$\exists~text_{j} \in Det_{C}$ \rm{\textbf{and}} $iou_{ik} > T$}{
        $m_{q}$ = \textbf{Overlap-Mask}($m_{i}$,$m_{k}$)\\
        $b_{q}$ = \textbf{Soft-Box}($b_{i}$,$b_{k}$)\\
        $w_{q}$ = $s_{i}*s_{k}*\alpha$\\
        $L_{p} = L_{p}\bigcup (m_{q},b_{q},w_{q})$
        }
        
    \Else{continue}
     }
    }
\end{algorithm}

Specifically, we first train three teacher models $A$, $B$, and $C$ with different data augmentations on labeled data, by which these models will focus on respective corresponding scenes and learn different text representations. Second, the three models perform multi-scale testing on unlabeled data to avoid missed detections as much as possible. Third, reliable pseudo labels for unlabeled data will be generated from the three sets of detection results through an ensemble strategy, which is described in Algorithm \ref{algorithm1}.

As shown, $Det_{A}$, $Det_{B}$, and $Det_{C}$ denote the detection result sets of model $A$, $B$, and $C$ respectively. We define a text instance as a triplet of $(m,b,s)$, which are the mask, bounding box, and score of the text. $L_{p}$ represents the final pseudo labels set including triplets of $(m,b,w)$, in which $w$ is the corresponding loss weight of proposals matched with this pseudo label during the training. For each text instance in $Det_{A}$, we retrieve the presence of text instances with high similarity in $Det_{B}$ and $Det_{C}$. If similar text instances appear in both $Det_{B}$ and $Det_{C}$, we consider these instances to be highly reliable, and then fuse them into a pseudo label with the multiplication of scores as $w$. If similar text instances exist only in $Det_{B}$ or $Det_{C}$, we consider these text instances to be weakly reliable and the $w$ of fused text instance will be decayed. In contrast, the text instance will be ignored when there is no similar text instance in $Det_{B}$ or $Det_{C}$. We calculate the Intersection over Union (IoU) score to evaluate the similarity between text instances. $T$ and $\alpha$ represent the IoU threshold and decay weight, and are set to 0.8 and 0.5 respectively. Soft-Box means the process of boxes merging by calculating the average of coordinates of boxes as the new coordinates of the pseudo labeled box. Meanwhile, we obtain the overlap of the text masks as the pseudo labeled mask in Overlap-Mask process. Finally, we train a student model on the combination of labeled data and pseudo labeled data.

Producing pseudo labels through multiple models mitigates the issue of missed detections caused by a single model with insufficient detection capability. Moreover, a pseudo label is jointly determined by multiple models, minimizing the problem of false positives and generating more accurate pseudo labels. The visualization example of the proposed pseudo labels generation method is shown in Figure \ref{fig8}.

\subsection{Loss Function}
Following Mask R-CNN, our model is trained in a multi-task manner, where a classification task, a box regression task, and an instance segmentation task are involved. Specifically, the final loss function is defined as follows:

\begin{equation}
\label{loss}
\quad\quad\quad\quad\quad\quad \textit{L} = \textit{L}_{\textit{rpn}} + \textit{L}_{\textit{box}} + \textit{L}_{\textit{mask}}
\end{equation}

where $L_{rpn}$ and $L_{box}$ denote the loss functions in the RPN and box branch, both of which consist of a Cross-Entropy ($Binary$ or $Softmax$) loss $L_{cls}$ and a Smooth L1 loss $L_{reg}$ for classification and box regression, respectively. $L_{mask}$ denotes the loss function in the mask branch, which is a Binary Cross-Entropy loss.

\section{Experiments}
In this section, we evaluate the performance of I3CL model on three public benchmarks, $i.e.$, ArT \citep{icdar2019art}, Total-Text \citep{totaltext}, and CTW-1500 \citep{ctw1500}, in which horizontal, quadrilateral, and curved texts exist simultaneously in most of the images. We first conduct comprehensive ablation studies to verify the effectiveness of proposed modules, and then compare I3CL with state-of-the-art methods.

\subsection{Datasets and Evaluation Metrics}
\textbf{SynthText} \citep{synthtext} is a dataset consisting of 800k synthetic images and 8 million text instances. We use it to pre-train our I3CL model.

\textbf{ArT} \citep{icdar2019art} is newly released dataset in the ICDAR2019 Robust Reading Challenge on Arbitrary-Shaped Text. It is the most challenging arbitrary-shaped text dataset containing Chinese texts, English texts, and other mixed symbols. It has a total of 10,166 images, including 5,603 training images and 4,563 testing images. Text instances in ArT are labeled by polygons with adaptive number of key points.

\textbf{Total-Text} \citep{totaltext} is a dataset that includes English texts of various shapes. It contains 1,255 images for training and 300 images for testing. All text instances are labeled by word-level polygons. 

\textbf{CTW-1500} \citep{ctw1500} is an English dataset focusing on curved texts, which consists of 1,000 training images and 500 testing images. Different from Total-Text, text instances in CTW-1500 are labeled by text-line-level polygons.

We follow the same standard evaluation protocols by using Recall, Precision, and F-measure as the evaluation metrics, which are provided by the dataset creators or competition organizers. F-measure is the major evaluation metric.

 \begin{figure*}[ht]
  \centering
  \includegraphics[width=0.95\linewidth]{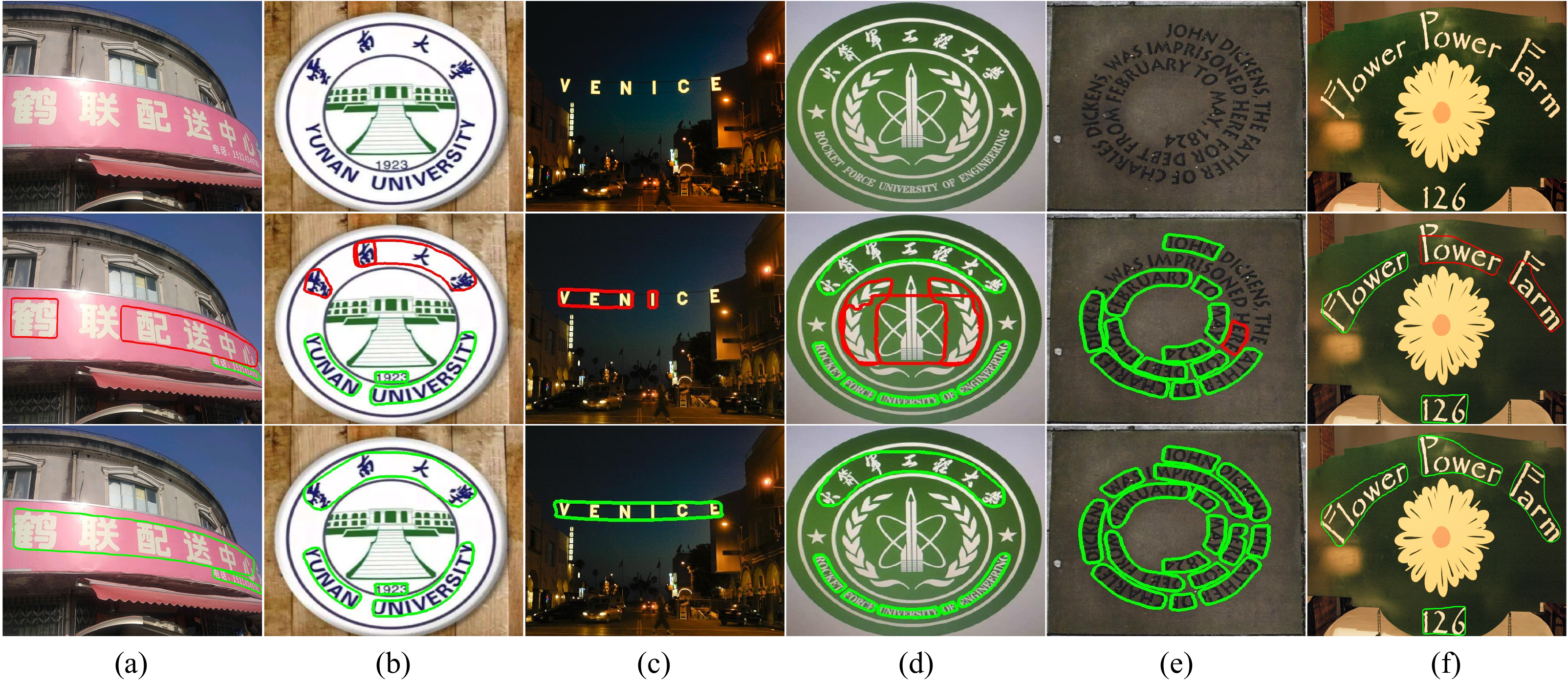}
  \caption{Detection results of Mask R-CNN (second row) and our I3CL model (third row). Mask R-CNN produces fracture detections(a-c), and inaccurate detections such as false positives(d), missed detections(e), as well as incomplete text contours(f), while our I3CL model can effectively mitigate these issues and generate more accurate detection results.}
  \label{fig9}
\end{figure*}

\subsection{Implementation Details}
We implement our proposed I3CL model based on the Detectron2 codebase with PyTorch. All experiments are performed using Nvidia Tesla V100 (16G) GPUs. The model is trained on 4 GPUs and tested on 1 GPU. As in the most of previous methods, we choose the ResNet-50 with the FPN as the backbone encoder.

\textbf{Training}. The whole training can be roughly divided into three main stages. Firstly, we pre-train a base model on SynthText dataset for about 300k iterations. Secondly, for each benchmark dataset, we fine-tune the base model using the corresponding real-world images for 30 epochs. In particular, considering that SynthText only contains English texts, we further pre-train the base model on the LSVT \citep{lsvt} and ICDAR2019-MLT \citep{icdar2019mlt} datasets before fine-tuning on ArT  to enhance the ability of the model regarding Chinese, following the common practice in \citep{crafts}. Finally, based on the proposed pseudo-label generation method, we choose the test data of LSVT as the unlabeled data for ArT, and make Total-Text and CTW-1500 act as unlabeled data for each other to further fine-tune the model in a semi-supervised learning way.

The batch size during pre-training is set to 8, and reduced to 4 during fine-tuning. We adopt the Stochastic gradient descent (SGD) optimizer to optimize our network with a weight decay of 0.0001 and a momentum of 0.9. During pre-training, the initial learning rate is set to 0.001 in the first 100k iterations and divided by 10 for the remaining iterations. For all experiments during fine-tuning, the initial learning rate is set to 0.0005 in the first 10 epochs and divided by 10 at 20 and 30 epochs. The shorter sides of images are randomly resized to different scales ($i.e.$, 800, 1,000, 1,200), and the upper limit of longer side is 1,800. Data augmentation strategies such as random noise, brightness adjusting, and color change are applied to increase the diversity of training data.

\textbf{Inference}. During inference, we only perform a sin-gle-scale test to evaluate our model. The shorter side of the test images was scaled to 1,000 while keeping the aspect ratio unchanged, and the maximum size of the longer side is limited to 2,100, 1,875, and 1,800 on three datasets. The function of $findContours$ in $OpenCV$ is used to generate polygon contours of text instances from predicted masks as the final detection results.

\subsection{Ablation Study}
\label{subsec:ablation}

\begin{table*}\normalsize
  \caption{Ablation study of the key components in our I3CL model on different datasets. ``Intra-CL'' represents the Intra-Instance Collaborative Learning module. ``Inter-CL(GCT)'' and ``Inter-CL(GCG)'' refer to Inter-Instance Collaborative Learning module with the global context structure based on transformer or global average pooling, respectively. ``R'', ``P'', and ``F'' represent Recall, Precision, and F-measure, respectively.}
  \label{tab1}
  \begin{center}
  \resizebox{0.9\hsize}{!}{
  \begin{tabular}{l|ccc|ccc|ccc|cc}
    \hline\hline
    \multirow{2}{*}{\textbf{Method}} & \multicolumn{3}{c|}{\textbf{ArT}} & \multicolumn{3}{c|}{\textbf{Total-Text}} & \multicolumn{3}{c|}{\textbf{CTW-1500}} & \multirow{2}{*}{\textbf{Parameter}} & \multirow{2}{*}{\textbf{GFLOPs}} \\
    \cline{2-10} & \textbf{R} & \textbf{P} & \textbf{F} & \textbf{R} & \textbf{P} & \textbf{F} & \textbf{R} & \textbf{P} & \textbf{F} \\
    \hline
    Baseline                            & 68.9 & 80.0 & 74.0  & 80.1 & 86.8 & 83.3  & 81.3 & 84.4 & 82.8 & 44.3M & 204.8\\
    Intra-CL                            & 69.4 & 81.7 & 75.0  & 82.2 & 87.5 & 84.7  & 82.8 & 85.5 & 84.1 & 46.7M & 245.6\\
    Intra-CL + Inter-CL(GCT)            & 70.9 & 82.8 & 76.4  & 83.4 & 88.8 & 86.0  & 84.4 & 87.3 & 85.8 & 58.5M & 248.8\\
    Intra-CL + Inter-CL(GCG)            & 71.3 & 82.7 & 76.6  & 83.7 & 89.2 & 86.3  & 84.5 & 87.4 & 85.9 & 52.2M & 247.3\\
    Intra-CL + Inter-CL(GCG\&GCT)        & 71.4 & 82.3 & 76.5  & 83.4 & 89.4 & 86.3  & 84.4 & 87.6 & 86.0 & 59.0M & 250.1\\
    Intra-CL + Inter-CL(GCG) + SSL & \textbf{72.2} & \textbf{83.6} & \textbf{77.5}  & \textbf{84.2} & \textbf{89.8} & \textbf{86.9}  & \textbf{84.6} & \textbf{88.4} & \textbf{86.5} & 52.2M & 247.3\\
  \hline\hline
\end{tabular}}
\vspace{-2mm}
\end{center}
\end{table*}

\begin{table*}
  \caption{Comparison results between the Mask R-CNN baseline and I3CL on three difficult subsets.}
  \label{tab2}
  \begin{center}
  \resizebox{0.9\hsize}{!}{
  \begin{tabular}{l|ccc|ccc|ccc}
    \hline\hline
    \multirow{2}{*}{\textbf{Method}} & \multicolumn{3}{c|}{\textbf{ArT subset}} & \multicolumn{3}{c|}{\textbf{Total-Text subset}} & \multicolumn{3}{c}{\textbf{CTW-1500 subset}}\\
    \cline{2-10} & \textbf{R} & \textbf{P} & \textbf{F} & \textbf{R} & \textbf{P} & \textbf{F} & \textbf{R} & \textbf{P} & \textbf{F} \\
    \hline
    Baseline        & 61.6 & 52.3 & 56.6  & 70.6 & 84.8 & 77.1  & 74.0 & 75.0 & 74.5 \\
    I3CL            & 68.8(\textcolor{blue}{$\uparrow$7.2}) & 61.2(\textcolor{blue}{$\uparrow$8.9}) & 64.8(\textcolor{blue}{$\uparrow$8.2})  & 80.0(\textcolor{blue}{$\uparrow$9.4}) & 89.8(\textcolor{blue}{$\uparrow$5.0}) & 84.6(\textcolor{blue}{$\uparrow$7.5})  & 82.1(\textcolor{blue}{$\uparrow$8.1}) & 82.6(\textcolor{blue}{$\uparrow$7.6}) & 82.4(\textcolor{blue}{$\uparrow$7.9}) \\
  \hline\hline
\end{tabular}}
\vspace{-2mm}
\end{center}
\end{table*}

\textbf{Effectiveness of each module}. We conduct an ablation study on ArT, Total-Text, and CTW-1500 to verify the effectiveness of each proposed module in this paper. For each dataset, we trained four models by adding the proposed modules gradually. ``\textbf{Baseline}'' denotes the original Mask R-CNN baseline model. ``\textbf{Int-ra-CL}'' denotes the model using the Intra-Instance Collaborative Learning module. ``\textbf{Inter-CL(GCT)}'' and ``\textbf{Inter-CL(GCG)}'' denote the model using the Inter-Instance Collaborative Learning module with the global context structure based on transformer or global average pooling, respectively. ``\textbf{SSL}'' refers to that I3CL model is trained in the semi-supervised learning way. The results are summarized in Table \ref{tab1}. 

As can be seen, \textbf{Intra-CL} improves the performance of the baseline model consistently on all three datasets, $e.g.$, 1.0\%, 1.4\%, and 1.3\% gains in terms of the F-measure on ArT, Total-Text, and CTW-1500, respectively. In addition, integrating it with \textbf{Inter-CL(GCT)} further brings absolute performance gains in terms of the F-measure increase by 1.4\%, 1.3\%, and 1.7\%, respectively. By contrast, the combination of \textbf{Intra-CL} and \textbf{Inter-CL(GCG)} achieves a better gain of 1.6\%, 1.6\%, and 1.8\% on F-measure but contains fewer parameters. In terms of F-measure, there is no significant gap between the two modules. The reason why GCG module works slightly better may be that each pixel of text RoI features integrates the complete global context. In terms of parameter, GCG module is much smaller than GCT module. Moreover, the combination of GCG module and GCT module in Inter-CL has no obvious advantage on F-measure and parameter. To this end, we choose the \textbf{Inter-CL(GCG)} as the default setting of Inter-CL module in I3CL. Finally, the I3CL model trained in a semi-supervised learning way achieves a gain of 3.5\%, 3.6\%, and 3.7\% in terms of the F-measure over the baseline on the three datasets, respectively. Moreover, there is a similar trend of improvement in the Precision and Recall. 

To prove that the improvements are obtained by addressing the two previous limitations, we select 100 images in each dataset, on which Mask R-CNN baseline is prone to produce a large number of fracture detections and inaccurate detections, as the difficult subset, and compare the performance gap between the Mask R-CNN baseline and the proposed I3CL on the three subsets. As shown in Table \ref{tab2}, I3CL achieves a remarkable gain of 8.2\%, 7.5\%, and 7.9\% in terms of F-measure on the three subsets respectively. Some visual results of the Mask R-CNN baseline and our I3CL model are shown in Figure \ref{fig9}. As can be seen, Mask R-CNN produces fracture detections, missed detections, as well as incomplete text contours indicated by the red boxes, while our I3CL model can effectively mitigate these issues and produce complete and accurate text masks. These quantitative and qualitative results demonstrate that our I3CL model benefits from the intra- and inter-instance collaborative learning and learns a better and more discriminative feature representation than Mask R-CNN baseline model, which help the text detector do well in detecting difficult texts.

\textbf{Different settings of the Intra-CL module}. Ablation studies are also conducted on the ArT dataset to investigate the impact of different settings of the Intra-CL module, $e.g.$, the number of convolutional branches in each block, feature fusion method, and the order of convolutional kernels in the cascade. The results are listed in Table \ref{tab2}. ``\textbf{1-path}'' denotes using a single convolutional branch with a regular $k \times k$ convolutional kernel. ``\textbf{2-path}'' denotes using two convolutional branches with asymmetric horizontal and vertical convolutional kernels. ``\textbf{3-path}'' denotes the default setting that contains all three branches as shown in Figure~\ref{fig5}. ``\textbf{cat}'' and ``\textbf{sum}'' denote the feature fusion method, $i.e.$, concatenation and element-wise sum, respectively. ``\textbf{kernel:753}'' denotes using the $7\times7$ convolutional kernel in the first block and $5\times5$ and $3\times3$ kernels subsequently.

\begin{table}
\begin{center}
  \caption{Ablation study of different settings of the Intra-CL module on the ArT dataset.}
   \vspace{-1mm}
  \label{tab3}
  \resizebox{0.9\hsize}{!}{
  \begin{tabular}{l|ccc}
    \hline\hline
    \textbf{Method} & \textbf{R} & \textbf{P} & \textbf{F}  \\
    \hline
    Baseline                                         & 68.9 & 80.0 & 74.0 \\
    Intra-CL (1-path)                                & 69.2 & 80.6 & 74.4 \\
    Intra-CL (2-path)                                & 69.1 & 80.9 & 74.5 \\
    Intra-CL (3-path, kernel:753, cat)               & 69.1 & \textbf{81.8} & 74.9 \\
    Intra-CL (3-path, kernel:753, sum)               & \textbf{69.4} & 81.7 & \textbf{75.0} \\
    Intra-CL (3-path, kernel:357, sum)               & 69.0 & 80.5 & 74.3\\
    \hline\hline
\end{tabular}}
 \vspace{-2mm}
\end{center}
\end{table}

\begin{table}
\begin{center}
  \caption{Ablation study on the setting of kernel sizes in the Intra-CL module on the ArT dataset.}
   \vspace{-1mm}
  \label{tab4}
  \resizebox{0.8\hsize}{!}{
  \begin{tabular}{l|ccc|c}
    \hline\hline
    \textbf{Method} & \textbf{R} & \textbf{P} & \textbf{F} & \textbf{Parameter} \\
    \hline
    kernel:753             & 69.4  & \textbf{81.7} & 75.0 & 46.7M \\
    kernel:975             & 69.3 & 81.6 & 75.0 & 48.20M\\
    kernel:777             & 69.6 & 80.9 & 74.8 & 48.10M\\
    kernel:555             & 69.5 & 81.0 & 74.8 & 46.4M\\
    kernel:333             & 69.5 & 80.7 & 74.7 & \textbf{45.1M}\\
    kernel:775             & 69.6 & 80.6 & 74.7 & 47.8M\\
    kernel:773             & \textbf{69.7} & 81.5 & \textbf{75.1} &  47.5M\\
    kernel:755             & 69.1 & 81.5 & 74.7 &  47.0M\\
    kernel:733             & 69.6 & 81.0 & 74.9 & 46.3M\\
    \hline\hline
\end{tabular}}
  \vspace{-2mm}
\end{center}
\end{table}

\begin{table}[ht]
  \caption{Ablation study of different settings of pseudo label generation on the ArT dataset.}
   \vspace{-1mm}
  \label{tab5}
  \begin{center}
  \resizebox{1\hsize}{!}{
  \begin{tabular}{l|ccc}
    \hline\hline
    \textbf{Method}  & \textbf{R} & \textbf{P} & \textbf{F}\\
    \hline
    Baseline                                        & 71.3 & 82.7 & 76.6 \\
    Threshold Filter                                 & 70.4 & 81.5 & 75.5 \\
    Ensemble                                        & 71.8 & 83.0 & 77.0 \\
    Ensemble + Overlap-Mask                           & 71.5 & \textbf{84.1} & 77.3 \\
    Ensemble + Overlap-Mask + Soft-Box                & \textbf{72.2} & 83.6 & \textbf{77.5} \\
  \hline\hline
\end{tabular}}
  \vspace{-2mm}
\end{center}
\end{table}

\begin{table}[ht]
  \caption{Evaluation results on the ArT dataset. "$\dag$", "$\ddag$", and "$\S$" indicate that the results are collected from \citep{icdar2019art}, official website of ArT, and  our experiments using official released code, respectively.}
  \vspace{-1mm}
  \label{tab6}
  \begin{center}
  \resizebox{1\hsize}{!}{
  \begin{tabular}{l|c|c|ccc}
    \hline\hline
    \textbf{Method} & \textbf{Venue} & \textbf{Backbone} & \textbf{R} & \textbf{P} & \textbf{F}\\
    \hline
    PSENet$\dag$ \citep{psenet}                & CVPR'19  & Res50       & 52.2 & 75.9 & 61.9 \\
    TextMountain $\dag$ \citep{textmountain}   & PR'21   & Res50   & 53.5 & \textbf{86.2} & 66.0 \\
    TextRay \citep{textray}                    & MM'20    & Res50       & 58.6 & 76.0 & 66.2 \\
    ContourNet$\S$ \citep{contournet}  & CVPR'20  & Res50  & 62.1 & 73.2 & 67.2 \\
    PAN$\S$ \citep{pan}       & CVPR'19  & Res18       & 61.1 & 79.4 & 69.1 \\
    CRAFT$\dag$ \citep{craft}                  & CVPR'19  & VGG16       & 68.9 & 77.2 & 72.9 \\
    PCR \citep{pcr}                            & CVPR'21  & DLA34       & 66.1 & 84.0 & 74.0 \\
    TextFuseNet$\ddag$ \citep{textfusenet}     & IJCAI'20 & Res50       & 69.4 & 82.6 & 75.4 \\
    \hline 
    \textbf{I3CL}                            & -    & Res50            & 71.3 & 82.7 & 76.6\\
    \textbf{I3CL + SSL}                      & -    & Res50            & \textbf{72.2} & 83.6 & \textbf{77.5}\\
  \hline\hline
\end{tabular}}
\vspace{-3mm}
\end{center}
\end{table}

As shown, although both ``\textbf{1-path}'' and ``\textbf{2-path}'' can improve the performance marginally, ``\textbf{3-path}'' bri-ngs more improvement over the baseline model in terms of the F-measure, $i.e.$, a gain of 1.0\%, confirming the value of using different kernels for modeling the multi-oriented texts. Besides, there is no significant difference between the concatenation and element-wise sum for feature fusion. We choose the latter one as the default setting. When the order of convolutional kernels in the cascade is reversed, $i.e.$, from ``\textbf{kernel:753}'' to ``\textbf{kernel:357}'', we observe a performance drop of 0.7\% in terms of the F-measure. We suspect the reason is that using a large kernel at the first block can effectively model long-range dependencies between characters, between gaps, and between characters and gaps, and using smaller kernels subsequently can gradually guide the Intra-CL module focus on the central region of the receptive field to learn a discriminative feature representation for either character or gap regions. However, when reversing the order, the Intra-CL module may struggle in extracting long-range context and be prone to noisy features in later blocks due to the large receptive fields. Besides, we conducted an ablation study on the setting of kernel sizes in the Intra-CL module. As shown in Table \ref{tab4}, compared with other settings, ``\textbf{kernel:753}'' achieves a better trade-off between performance and parameters. Although ``\textbf{kernel:773}'' delivers slightly better results in terms of recall and F-measure, its precision score is worse than that of ``\textbf{kernel:753}'' while increasing the number of parameters by 0.8M. As a result, we use ``\textbf{kernel:753}'' as the default setting in the Intra-CL module.

\textbf{Different settings of pseudo label generation}. Moreover, we also conduct an ablation study to compare different settings of pseudo label generation on the ArT dataset. ``\textbf{Threshold Filter}'' denotes the common practice that selects the detection results from a single model as pseudo labels via a fixed threshold of confidence score. ``\textbf{Ensemble}'' represents the ensemble strategy based on multiple models without Inter-Mask and Soft-Box to refine the masks and boxes respectively. We use ``\textbf{Ensemble + Overlap-Mask + Soft-Box}'' as the default setting in our pseudo label generation. As shown in Table \ref{tab5}, due to the missed and false detections, ``\textbf{Threshold Filter}'' results in a severe side effect on the performance of the baseline model, $i.e.$ 0.9\%, 1.2\%, and 1.1\% drop on the three metrics respectively. In contrast, ``\textbf{Ensemble + Overlap-Mask}'' achieves the highest Precision of 84.1\% among all models with ResNet-50 backbone on ArT. Furthermore, the default ``\textbf{Ensemble + Overlap-Mask + Soft-Box}'' can bring considerable performance gains of 0.9\%, 0.9\%, and 0.9\% on the three metrics respectively, which verifies the effectiveness of our method.

\subsection{Comparison with State-of-the-art Methods}

\noindent\textbf{Evaluation on ArT}. We evaluate the effectiveness of the proposed I3CL model in detecting arbitrary-shaped mixed-language text on ArT dataset. The evaluation results of I3CL and previous methods are presented in Table \ref{tab6}. I3CL has achieved the best performance in terms of Recall and F-measure without using semi-supervised learning, which surpasses the current best method, $i.e.$, TextFuseNet$\dag$ \citep{textfusenet}, by a large margin of 1.2\% in terms of the F-measure. When applying the semi-supervised learning, a more compelling result can be achieved, $i.e.$, 72.2\% on Recall and 77.5\% on F-measure. To the best of our knowledge, I3CL is the first method achieving an F-measure over 77.0\% on ArT using the ResNet-50 backbone.

\noindent\textbf{Evaluation on Total-Text}. We evaluate the effectiveness of the proposed I3CL model in detecting word-level arbitrary-shaped English text on Total-Text dataset. As shown in Table \ref{tab7}, similarly, I3CL sets a new state-of-the-art result of 86.3\% F-measure on Total-Text. Furthermore, our detector with full implementations outperforms the latest state-of-the-art method, $i.e.$, FCEN-et \citep{fcenet}, by 1.1\% F-measure. Moreover, I3CL achieves the highest Precision of 89.8\%, which has a gain of 0.5\% over the previous best method. In addition, our I3CL is the only one exceeding 86.0\% in terms of the F-measure in all contenders.

\noindent\textbf{Evaluation on CTW-1500}. We evaluate the effectiveness of the proposed I3CL model in detecting text-line-level arbitrary-shaped English text on CTW-1500 dataset. The results are summarized in Table \ref{tab8}. As can be seen, our model achieves the best results with Precision of 88.4\% and F-measure of 86.5\% while keeping highly competitive results on Recall. Compared with the previous best method FCENet \citep{fcenet}, I3CL outperforms it by 1.2\% on Recall, 0.8\% on Precision, and 1.0\% on F-measure. Note that since the text instances in CTW-1500 are labelled by text-line-level polygons rather than word-level annotations in Total-text, our method can learn better local and long-range features to handle such challenging cases and produce more accurate detections.

\subsection{Competition on ArT Leaderboard}
To explore the upper limit of the detection performance of I3CL, we join the melee on the ArT leaderboard. More complex experiments are conducted to improve the performance of I3CL on ArT dataset, including using stronger backbones and other common tricks during the training and testing.

\textbf{Backbone}. We adopt ResNet \citep{resnet} and its variants, $i.e.$, ResNeXt \citep{resnext} and ResNeSt \citep{resnest}, with different depths of \{50,101,152\} as the backbone. Besides, we also adopt ResNet-50 pretrained with RegionCL \citep{regioncl} pretext task and transformer-based ViTAEv2-S \citep{zhang2022vitaev2} backbone for further experiments. The comparisons of different backbones can be seen in Table \ref{tab9}. As shown, RegionCL contrastive learning on ResNet-50 backbone apparently assists the downstream scene text detection task. Due to the superior feature representation by incorporating transformers with intrinsic inductive bias, ViTAEv2-S surpasses the base ResNet-50 with a large margin using similar parameters, $i.e.$, 2.3\% gain on F-measure. What's more, deeper backbones bring effective gains on all three evaluation metrics compared to the base ResNet-50. Among all the backbones we have tried, ResNeSt-101 stands out with the highest values on Recall, Precision, and F-measure at single-scale testing, $i.e.$, 75.1\%, 86.3\%, and 80.3\%, which are even better than the deeper ResNeXt-152. After comprehensive consideration about the size, training speed, and memory consumption of the model, we chose the ResNeSt-101 as the final backbone.

\begin{figure*}[ht]
  \centering
  \includegraphics[width=\linewidth]{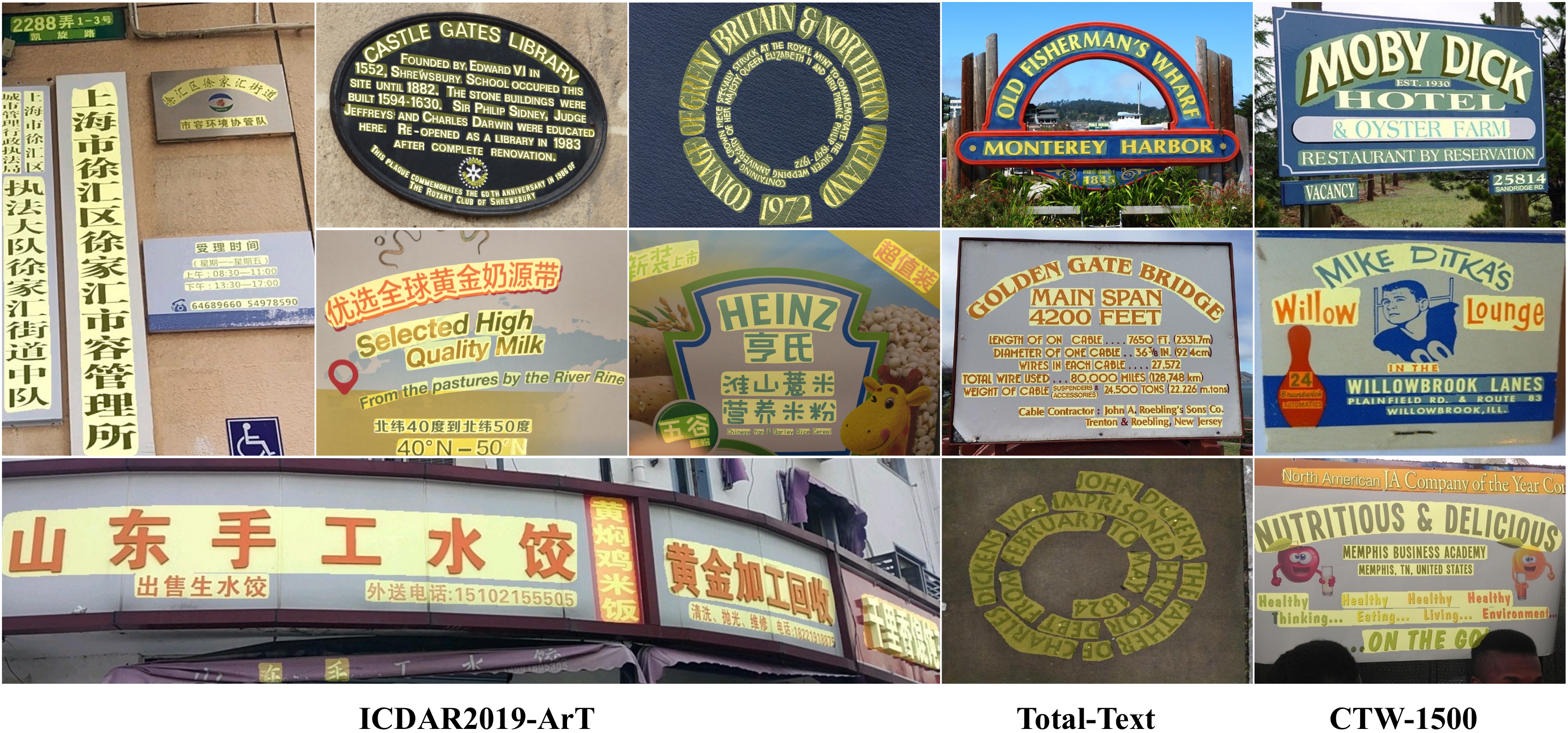}
  \caption{Some visual results of our I3CL model on the ArT, Total-Text, and CTW-1500 datasets, respectively.}
  \label{fig10}
\end{figure*}

\begin{table}[ht]
  \caption{Evaluation results on the Total-Text dataset.}
  \label{tab7}
  \begin{center}
  \resizebox{1\hsize}{!}{
  \begin{tabular}{l|c|c|ccc}
    \hline\hline
    \textbf{Method} & \textbf{Venue} & \textbf{Backbone} & \textbf{R} & \textbf{P} & \textbf{F}\\
    \hline
    Mask-TTD \citep{mask-ttd}           & TIP'19       & Res50              & 74.5 & 79.1 & 76.7 \\
    TextSnake \citep{textsnake}         & ECCV'18      & VGG16              & 74.5 & 82.7 & 78.4 \\
    ATRR \citep{atrr}                   & CVPR'19      & VGG16              & 76.2 & 80.9 & 78.5 \\
    MSR \citep{msr}                     & IJCAI'19     & Res50              & 74.8 & 83.8 & 79.0 \\
    CSE \citep{cse}                     & CVPR'19      & Res34              & 79.1 & 81.4 & 80.2 \\
    SAST \citep{sast}                   & MM'19        & Res50              & 76.9 & 83.8 & 80.2 \\
    TextDragon \citep{textdragon}       & ICCV'19      & VGG16              & 75.7 & 85.6 & 80.3 \\
    TextRay \citep{textray}             & MM'20        & Res50              & 77.9 & 83.5 & 80.6 \\
    TextField \citep{textfield}         & TIP'19       & VGG16              & 79.9 & 81.2 & 80.6 \\
    PSENet \citep{psenet}               & CVPR'19      & Res50              & 78.0 & 84.0 & 80.9 \\
    SegLink++ \cite{seglink++}           & PR'19        & VGG16              & 80.9 & 82.1 & 81.5 \\
    MS-CAFA \citep{mscafa}              & TMM'19       & Res50              & 78.6 & 84.6 & 81.5 \\
    SPCNet \citep{spcnet}               & AAAI'19      & Res50              & 82.8 & 83.0 & 82.9 \\
    LOMO \citep{lomo}                   & CVPR'19      & Res50              & 79.3 & 87.6 & 83.3 \\
    CRAFT \citep{craft}                 & CVPR'19      & VGG16              & 79.9 & 87.6 & 83.6 \\
    CRNet \citep{crnet}                 & MM'20        & Res50              & 82.5 & 85.8 & 84.1 \\
    Boundary \citep{boundary}           & AAAI'20      & Res50              & 83.5 & 85.2 & 84.3 \\
    ABCNet \citep{abcnet}               & CVPR'20      & Res50              & 81.3 & 87.9 & 84.5 \\
    DB \citep{db}                       & AAAI'20      & Res50-DCN          & 82.5 & 87.1 & 84.7 \\
    PAN \citep{pan}                     & ICCV'19      & Res18              & 81.0 & 89.3 & 85.0 \\
    TextPerception \citep{textperceptron} & AAAI'20    & Res50              & 81.8 & 88.8 & 85.2  \\
    PCR \citep{pcr}                     & CVPR'21      & DLA34              & 82.0 & 88.5 & 85.2  \\
    TextFuseNet \citep{textfusenet}     & IJCAI'20     & Res50              & 83.2 & 87.5 & 85.3 \\
    ContourNet \citep{contournet}       & CVPR'20      & Res50              & 83.9 & 86.9 & 85.4 \\
    DGGR \citep{dggrn}                  & CVPR'20      & VGG16              & \textbf{84.9} & 86.5 & 85.7\\
    FCENet \citep{fcenet}               & CVPR'21      & Res50-DCN          & 82.5 & 89.3 & 85.8 \\
    \hline
    \textbf{I3CL}                      & -              & Res50              & 83.7 & 89.2 & 86.3\\
    \textbf{I3CL + SSL}                & -              & Res50              & 84.2 & \textbf{89.8} & \textbf{86.9}\\
  \hline\hline
\end{tabular}}
\vspace{-3mm}
\end{center}
\end{table}

\begin{table}[ht]
  \caption{Evaluation results on the CTW-1500 dataset.}
  \label{tab8}
  \begin{center}
  \resizebox{1\hsize}{!}{
  \begin{tabular}{l|c|c|ccc}
    \hline\hline
    \textbf{Method} & \textbf{Venue} & \textbf{Backbone} & \textbf{R} & \textbf{P} & \textbf{F}  \\
    \hline
    CTD \citep{ctd}                       & PR'19       & Res50          & 69.8 & 77.4 & 73.4 \\
    TextSnake \citep{textsnake}           & ECCV'18     & VGG16          & \textbf{85.3} & 67.9 & 75.6 \\
    CSE \citep{cse}                       & CVPR'19     & Res34          & 76.0 & 81.1 & 78.4 \\
    Mask-TTD \citep{mask-ttd}             & TIP'19      & Res50          & 79.0 & 79.7 & 79.4 \\
    ATRR \citep{atrr}                     & CVPR'19     & VGG16          & 80.2 & 80.1 & 80.1 \\
    SAE \citep{sae}                       & CVPR'19     & Res50          & 77.8 & 82.7 & 80.1 \\
    LOMO \citep{lomo}                     & CVPR'19     & Res50          & 76.5 & 85.7 & 80.8 \\
    SAST \citep{sast}                     & MM'19       & Res50          & 77.1 & 85.3 & 81.0 \\
    SegLink++ \citep{seglink++}           & PR'19       & VGG16          & 79.8 & 82.8 & 81.3 \\
    TextField \citep{textfield}           & TIP'19      & VGG16          & 79.8 & 83.0 & 81.4 \\
    ABCNet \citep{abcnet}                 & CVPR'20     & Res50          & 78.5 & 84.4 & 81.4 \\
    MSR  \citep{msr}                      & IJCAI'19    & Res50          & 78.3 & 85.0 & 81.5 \\
    TextRay \citep{textray}               & MM'20       & Res50          & 80.4 & 82.8 & 81.6 \\
    PSENet \citep{psenet}                 & CVPR'19     & Res50          & 79.7 & 84.8 & 82.2 \\
    DB \citep{db}                         & AAAI'20     & Res50-DCN      & 80.2 & 86.9 & 83.4 \\
    CRAFT \citep{craft}                   & CVPR'19     & VGG16          & 81.1 & 86.0 & 83.5 \\
    TextDragon \citep{textdragon}         & ICCV'19     & VGG16          & 82.8 & 84.5 & 83.6 \\
    PAN \citep{pan}                       & ICCV'19     & Res18          & 81.2 & 86.4 & 83.7 \\
    CRNet \citep{crnet}                   & MM'20       & Res50          & 80.9 & 87.0 & 83.8 \\
    ContourNet \citep{contournet}         & CVPR'20     & Res50          & 84.1 & 83.7 & 83.9 \\
    DGGR \citep{dggrn}                    & CVPR'20     & VGG16          & 83.0 & 85.9 & 84.5 \\
    TextPerception \citep{textperceptron} & AAAI'20     & Res50          & 81.9 & 87.5 & 84.6 \\
    PCR \citep{pcr}                       & CVPR'21     & DLA34          & 82.3 & 87.2 & 84.7 \\
    TextFuseNet \citep{textfusenet}       & IJCAI'20    & Res50          & 85.0 & 85.8 & 85.4 \\
    FCENet \citep{fcenet}                 & CVPR'21     & Res50-DCN      & 83.4 & 87.6 & 85.5 \\
    \hline
    \textbf{I3CL}                     & -                & Res50          & 84.5 & 87.4 & 85.9\\
    \textbf{I3CL + SSL}               & -                & Res50          & 84.6 & \textbf{88.4} & \textbf{86.5}\\
  \hline\hline
\end{tabular}}
\end{center}
\end{table}

\textbf{Multi-scale Testing}. We resized the maximum size of the longer side to \{1,500, 1,800, 2,100, 2,400, 2,700\}, and the shorter side is scaled to \{1,000, 1,300, 1,500\}. Results from different scales are aggregated and NMS is used to suppress the redundant text instances to get final detection results. As can be seen in Table \ref{tab8}, compared to single-scale testing, multi-scale testing achieves a margin of 0.9\% on F-measure over the baseline.

\textbf{Soft-NMS} \citep{softnms}. NMS is replaced by Soft-NMS during the testing. We tried two decay strategies of confidence score, including linear decay and Gaussian decay. A finding in our experiment is that Soft-NMS with linear decay is better for ArT, and the F-measure was increased from 81.2\% to 82.0\%.

\textbf{Mixup} \citep{mixup}. We implemented Mix-up by pasting two text images and their labels in a fixed transparency ratio, $i.e.$, $1$:$1$. Besides, we paste text images and non-text images together in a random transparency ratio. Those non-text images include landscape, architectural and animal images, increasing the diversity of background. In this way, I3CL achieves 80.8\% in terms of F-measure at single-scale testing without Soft-NMS and 82.3\% at multi-scale testing with Soft-NMS.

\textbf{Model Ensemble}. We ensemble the detected text boxes of different models to obtain better final results, such as models training with different backbones, different data augmentation strategies, and different iterations. Similarly in multi-scale testing, detection results from models using different training strategies are aggregated and then we use Soft-NMS to remove redundant text instances. As shown in Table \ref{tab8}, our I3CL ultimately achieves an extremely impressive F-measure of 84.0\% on ArT dataset.

In summary, the proposed I3CL sets new state-of-the-art on the ArT, Total-Text, and CTW-1500 for arbitrary-shaped scene text detection. Specifically, I3CL with ResNeSt-101 backbone achieves an impressive detection performance and ranks the $1^{st}$ place on the ArT leaderboard. Some visual results are shown in Figure \ref{fig10}. As can be seen, I3CL can well handle different challenging cases including various shapes, extremely small scales, large gaps between characters, diverse font styles, and backgrounds, showing great potential for real-world applications.

\begin{table}[ht]\scriptsize
  \caption{Results of I3CL without SSL using different backbones on ArT dataset. $\dag$ and $\ddag$ represent the RegionCL~\citep{regioncl} with finetuning and without finetuning on the ImageNet training data, respectively. * indicates that the whole detection model is implemented in MMDetection \citep{mmdetection}. }
  \label{tab9}
  \begin{center}
  \resizebox{0.7\hsize}{!}{
  \begin{tabular}{l|ccc}
    \hline\hline
    \textbf{Backbone}  & \textbf{R} & \textbf{P} & \textbf{F}\\
    \hline
    ResNet-50              & 71.3 & 82.7 & 76.6 \\
    ResNet-50 w/ RegionCL$\dag$     & 72.6 & 81.9 & 77.0 \\
    ResNet-50 w/ RegionCL$\ddag$     & 73.5 & 81.6 & 77.3 \\
    ViTAEv2-S*                 & 75.4 & 82.8 & 78.9 \\
    ResNeXt-101            & 74.1 & 85.5 & 79.4 \\
    ResNeSt-101            & \textbf{75.1} & \textbf{86.3} & \textbf{80.3} \\
    ResNeXt-152            & 74.9 & 86.0 & 80.1 \\
  \hline\hline
\end{tabular}}
\end{center}
\end{table}

\begin{table}[ht]
  \caption{Results of I3CL using different tricks with ResNeSt-101 backbone on ArT dataset.}
  \label{tab8}
  \begin{center}
  \resizebox{1\hsize}{!}{
  \begin{tabular}{cccc|c}
    \hline\hline
    \textbf{MS Testing}  & \textbf{Soft-NMS} & \textbf{Mixup} & \textbf{Model Ensemble} &\textbf{F}\\
    \hline
        &   &   &   & 80.3 \\
    \checkmark    &   &   &   & 81.2 \\
    \checkmark    &\checkmark   &   &   & 82.0 \\
    \checkmark    &\checkmark   &\checkmark   &   & 82.3 \\
    \checkmark    &\checkmark   &\checkmark   &\checkmark   & \textbf{84.0} \\
  \hline\hline
\end{tabular}}
\end{center}
\end{table}

\section{Discussion about Model Complexity}
In this section, we discuss the model complexity of our I3CL, including parameter, computation, and inference speed.

\textbf{Parameter}. As shown in Table \ref{tab1}, I3CL brings considerable performance improvements of over 3\% F-mea-sure on different datasets with 17.8\% parameters increase. The main parameter increase comes from transformer in Inter-CL module. ContourNet \citep{contournet} using Deformable ROI pooling with 256.3M parameters achieved 67.2\%, 85.4\%, and 83.9\% in terms of F-measure on ArT, Total-Text, and CTW-1500, respectively. In contrast, I3CL sets new state-of-the-art results with 77.5\%, 86.9\%, and 86.5\% on the three data-sets and maintains a better trade-off between the model size and the performance with only 52.2M parameters, which proves that effectively model design for specific problems is important.

\textbf{Computation}. As known, transformer often brings heavy computation due to the self-attention mechanism. Unlike the image classification task in \citep{xu2021vitae}, \citep{zhang2022vitaev2} and \citep{swin}, the computation of transformer encoder for modeling the dependencies between text instances in Inter-CL module depends on the number of text instances and the dimension of sequence features. After statistics, the average number of text instances on the images of the three datasets is 9. Benefit by the dimension reduction of the input sequence features and reasonable depth of transformer structure, transformer in Inter-CL module only increases about 0.05 GFLOPs computation on average. Compare with the 204.8 GFLOPs computation of the Mask R-CNN baseline, we consider that the computation increases of transformer structure when modeling the dependencies between text instances in Inter-CL module is acceptable. The total computation of I3CL can be seen in Table \ref{tab1}. Overall, I3CL brings considerable performance improvements of over 3\% F-measure on different datasets with 20.7\% computation increase.

\textbf{Inference Speed}. I3CL achieves an inference speed of 7.6 fps on CTW-1500 dataset, which is slightly slower than 9.1 fps of the Mask R-CNN baseline. The comparison results of speed between I3CL and some previous methods can be seen in Table \ref{tab10}. When testing on CTW-1500, I3CL surpasses PSENet \citep{psenet} and ContourNet \citep{contournet} both on F-measure and speed ($i.e.$, 86.5\% $vs$ 82.2\% and 83.9\% and 7.6 fps $vs$ 3.9 fps and 4.5 fps). However, compared with DB \citep{db}, though I3CL outperforms it by a large margin on F-measure ($i.e.$, 86.5\% $vs$ 83.5\%), our method lags behind on speed ($i.e.$, 7.6 fps $vs$ 22.0 fps). Overall, Although the inability to achieve real-time detection is a congenital limitation of the two-stage detector, I3CL still has obvious advantages over some previous methods on inference speed.

\begin{table}[ht]\scriptsize
  \caption{Comparision results of speed between I3CL and some previous methods on CTW-1500 dataset.}
  \label{tab10}
  \begin{center}
  \resizebox{1.0\hsize}{!}{
  \begin{tabular}{l|c|c|c}
    \hline\hline
    \textbf{Method} & \textbf{Venue} & \textbf{Backbone} & \textbf{FPS}\\
    \hline
    CSE \citep{cse}                 & CVPR'19  & Res34 & 2.6 \\
    PSENet \citep{psenet}           & CVPR'19  & Res50 & 3.9 \\
    MSR \citep{msr}                 & IJCAI'19 & Res50 & 4.3 \\
    LOMO \citep{lomo}               & CVPR'19  & Res50 & 4.4 \\
    ContourNet \citep{contournet}   & CVPR'20  & Res50 & 4.5 \\
    TextField \citep{textfield}     & TIP'19   & VGG16 & 6.0 \\
    TextFuseNet \citep{textfusenet} & IJCAI'20 & Res50 & 7.3 \\
    PCR \citep{pcr}                 & CVPR'21  & DLA34 & 11.8 \\
    DB \citep{db}                   & AAAI'20  & Res50 & \textbf{22.0} \\
    \hline
    \textbf{I3CL}                           & -  & Res50 & 7.6 \\
  \hline\hline
\end{tabular}}
\end{center}
\end{table}

\begin{figure}[ht]
  \centering
  \includegraphics[width=0.95\linewidth]{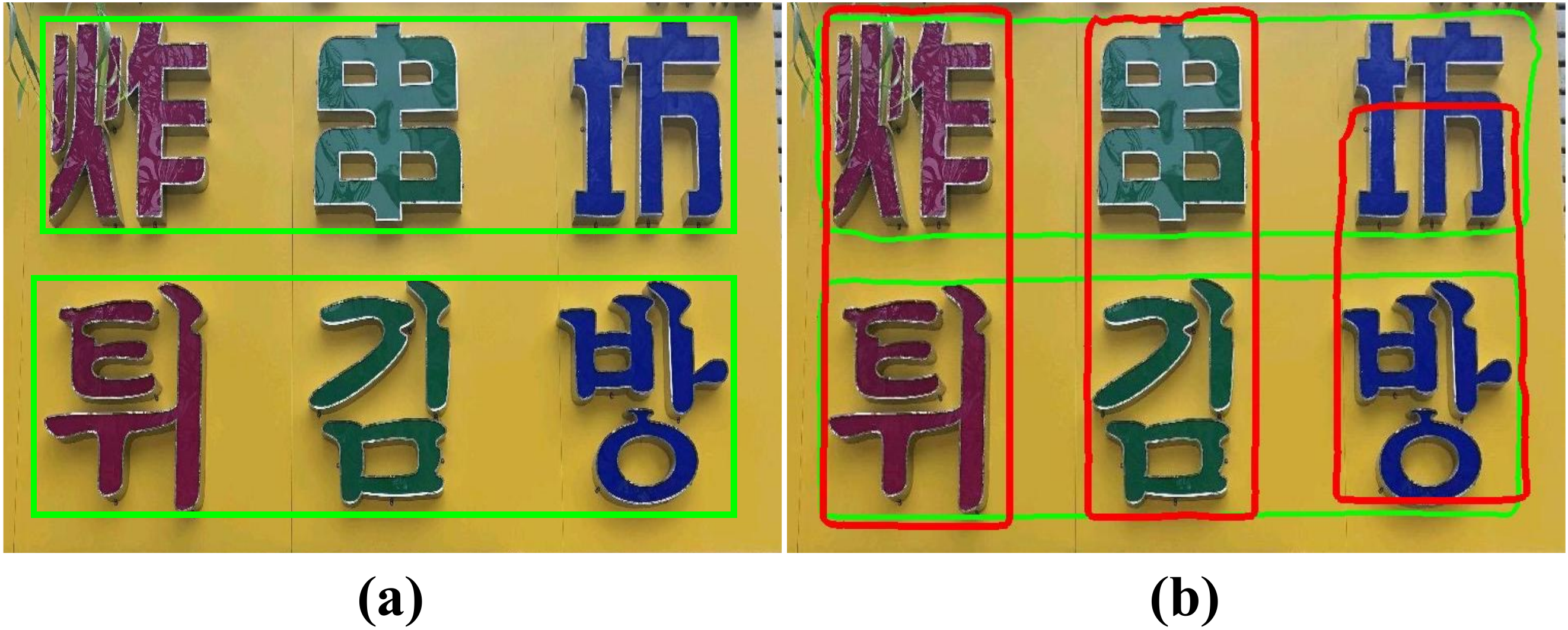}
  \caption{(a) A test image with two horizontal text regions. (b) Failure detection. In (b), green polygons represent true positives, while red polygons represent false positives.}
  \label{fig11}
\end{figure}

\section{Limitation}
In this section, we discuss the limitations of the proposed I3CL model. Although achieving state-of-the-art performance on three challenging benchmarks, our method is not outstanding enough in terms of speed, which can not meet the requirement for real-time applications. In the future, we plan to investigate efficient instance segmentation pipelines and fast implementation as well as other effective and lightweight modules for collaborative learning. In addition, our model may generate linguistically ambiguous text proposals when detecting text arranged in multiple rows and columns. A failure detection example is shown in Figure \ref{fig11}. In the future, domain knowledge of linguistics can be utilized to design more effective modules as well as grouping strategies for proposal generation and filtering to mitigate the issue.

\section{Conclusion}
In this paper, we first identify two issues in arbitrary-shaped text detection, $i.e.$, fracture detection and inaccurate detection, and then argue that collaborative learning of both character and gap regions in text and long-dependencies between text instances within an image matters for mitigating the two issues. To validate the idea, we make the first attempt to propose a novel intra- and inter-instance collaborative learning model named I3CL, where an Intra-CL module based on a cascade of convolutional blocks with multiple receptive fields and an Inter-CL module based on a text instance transformer are devised. Besides, a new method of pseudo label generation based on ensemble strategy is proposed for semi-supervised learning of scene text detection. Comprehensive empirical studies on three public benchmarks demonstrate the effectiveness of the proposed I3CL model and its superiority over existing methods. We hope this study can open a new perspective for text detection and encourage more follow-up work in modeling long-range dependencies within and between text instances.

%
%

\begin{acknowledgements}
This work was supported in part by National Natural Science Foundation of China: Grant No. 62076186, 62141112 and 41871243, in part by Science and Technology Major Project of Hubei Province (Next-Generation AI Technologies): Grant No. 2019AEA170, and in part by ARC FL-170100117. The numerical calculations in this paper have been done on the supercomputing system in the Supercomputing Center of Wuhan University.
\end{acknowledgements}

%
%

\bibliographystyle{spbasic}      
\bibliography{reference.bib}   


\end{document}